\documentclass[WileyNJD-MPS,Times2COL]{WileyNJDv5} 

\hypersetup{colorlinks, anchorcolor=black, 
            linkcolor=blue, citecolor=blue}
\usepackage{bbding}
\usepackage{graphicx}
\usepackage{makecell}

\articletype{ORIGINAL RESEARCH}%

\received{Date Month Year}
\revised{Date Month Year}
\accepted{Date Month Year}
\journal{CAAITrans.Intell.Technol}
\volume{xx}
\copyyear{xxxx}
\startpage{1}

\begin{document}

\title{Explore Human Parsing Modality for Action Recognition}

\author[1]{Jinfu Liu$^*$}

\author[2]{Runwei Ding$^*$}

\author[1]{Yuhang Wen}

\author[4]{Nan Dai}

\author[3]{Fanyang Meng}

\author[1]{Shen Zhao$^\dagger$}

\author[2]{Mengyuan Liu$^\dagger$}

\authormark{Liu \& Ding \textsc{et al.}}
\titlemark{Explore Human Parsing Modality for Action Recognition}

\address[1]{\orgdiv{School of Intelligent Systems Engineering}, \orgname{Sun
Yat-sen University}, \orgaddress{\state{Shenzhen}, \country{China}}}

\address[2]{\orgdiv{The Key Laboratory of Machine Perception, Shenzhen Graduate School}, \orgname{Peking University}, \orgaddress{\state{Shenzhen}, \country{China}}}

\address[3]{\orgname{Peng Cheng Laboratory}, \orgaddress{\state{Shenzhen}, \country{China}}}

\address[4]{\orgname{Changchun University of Science and Technology}, \orgaddress{\state{Changchun}, \country{China}}}

\corres{Shen Zhao \email{zhaosh35@mail.sysu.edu.cn} \\
        Mengyuan Liu \email{nkliuyifang@gmail.com} \\
        $^*$ means co-first authors.}


\abstract[Abstract]{Multimodal-based action recognition methods have achieved high success using pose and RGB modality. However, skeletons sequences lack appearance depiction and RGB images suffer irrelevant noise due to modality limitations. To address this, we introduce human parsing feature map as a novel modality, since it can selectively retain effective semantic features of the body parts, while filtering out most irrelevant noise. We propose a new dual-branch framework called Ensemble Human Parsing and Pose Network (EPP-Net), which is the first to leverage both skeletons and human parsing modalities for action recognition. The first human pose branch feeds robust skeletons in graph convolutional network to model pose features, while the second human parsing branch also leverages depictive parsing feature maps to model parsing festures via convolutional backbones. The two high-level features will be effectively combined through a late fusion strategy for better action recognition. Extensive experiments on NTU RGB+D and NTU RGB+D 120 benchmarks consistently verify the effectiveness of our proposed EPP-Net, which outperforms the existing action recognition methods. Our code is available at: \href{https://github.com/liujf69/EPP-Net-Action}{\textcolor{black}{https://github.com/liujf69/EPP-Net-Action}}.}



\maketitle

\renewcommand\thefootnote{}

\renewcommand\thefootnote{\fnsymbol{footnote}}
\setcounter{footnote}{1}

\section{INTRODUCTION}\label{sec1}
Human action recognition is an important task in the field of computer vision, which also has great research value and broad application prospects in education\cite{9383294}, human-computer interaction\cite{7102751} and content-based video retrieval\cite{7858791}. It can also be integrated with fields such as motion prediction\cite{wang2024dynamic}, pose estimation\cite{wang2023global} and micro-expression generation\cite{zhang2024facial} to accomplish more complex human-related tasks. Most related methods\cite{CTR-GCN2021,shiftgcn2020,dynamicgcn2020,zhang2022zoom} use unimodal data for action recognition, typically skeleton-based action recognition, which take human skeletons\cite{liu2023novel,liu2022generalized,wen2023interactive,STTFormer} as the input. In recent years, methods\cite{vpn2020,9680676} of integrating multiple modalities have emerged to make effective use of multimodal features for better action recognition. Among them, the human skeleton and RGB modality are two widely-adopted input modalities.

Actually the skeleton modality can be viewed as a natural topological graph, where the graph vertices and edges represent joints and bones of human body respectively. The graph-structured skeleton can well represent body movements and is highly robust to environmental changes\cite{survey}, thereby adopted in many studies using graph convolutional networks (GCNs)\cite{tu2022joint,shiftgcn2020,CTR-GCN2021,Chi_2022_CVPR,SATD,wang2024dynamic}. Besides, the RGB modality has rich appearance information, therefore some prior studies in action recognition employ convolutional neural networks (CNNs)\cite{vpn2020,tu2019action} to model spatiotemporal features from RGB images and achieve the desired effect. In multimodal-based action recognition, the features and complementarity between modalities are vital. However, most existing methods\cite{vpn2020,9680676} based on skeleton and RGB modality have some limitations in representation due to the input modality. For instance, the skeleton modality lacks the ability to depict the appearance of human body parts, while the RGB modality is prone to being influenced by various sources of noise, including background interference and changes in illumination. \uline{So it is meaningful to explore another modality that incorporates body-part appearance depiction while remaining noiseless and robust}.

\begin{figure}[t]
\centerline{\includegraphics[width=8cm]{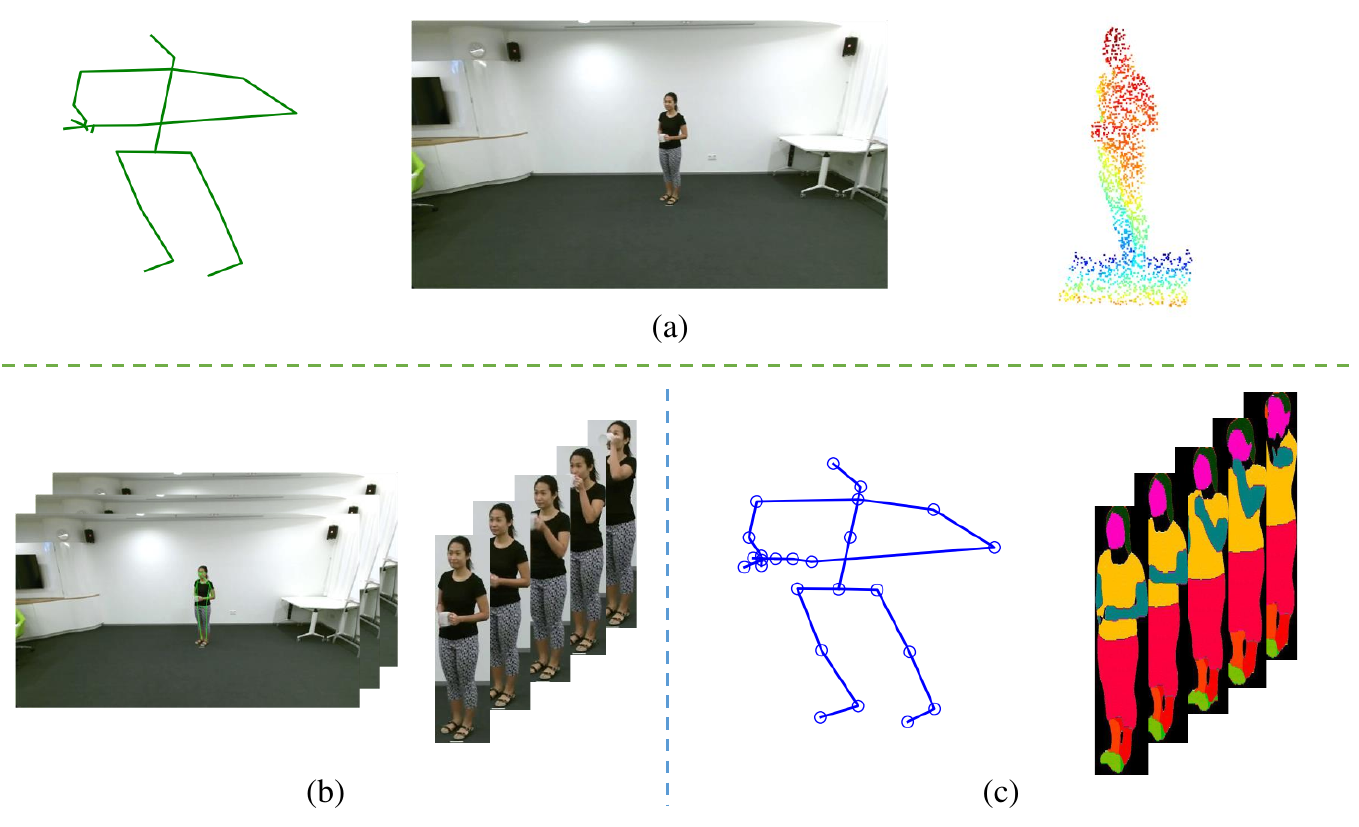}}
\caption{Action recognition using different modalities. (a) Previous methods using unimodal modality like skeleton, RGB or point cloud data. (b) Previous multimodal methods using noisy RGB modality. (c) Our EPP-Net is the first to introduce novel human parsing modality and leverage both skeletons and human parsing feature maps for multimodal-based action recognition.}
\label{modality}
\end{figure}

Inspired by this motivation, we intend to introduce the human parsing into action recognition. Actually the human parsing facilitates the recognition of distinct semantic parts\cite{LIP2019} and effectively eliminates irrelevant details while retaining crucial extrinsic features of the human body. Meanwhile, we propose an Ensemble Human Parsing and Pose Network (EPP-Net) for multimodal-based action recognition. Unlike previous methods using unimodal modality (Fig. \ref{modality} (a)) or using 
noisy RGB modality (Fig. \ref{modality} (b)) in multimodal learning, our EPP-Net (Fig. \ref{modality} (c)) is the first to combine human parsing and pose modality for action recognition.

\begin{figure*}[t]
\centerline{\includegraphics[width=17cm]{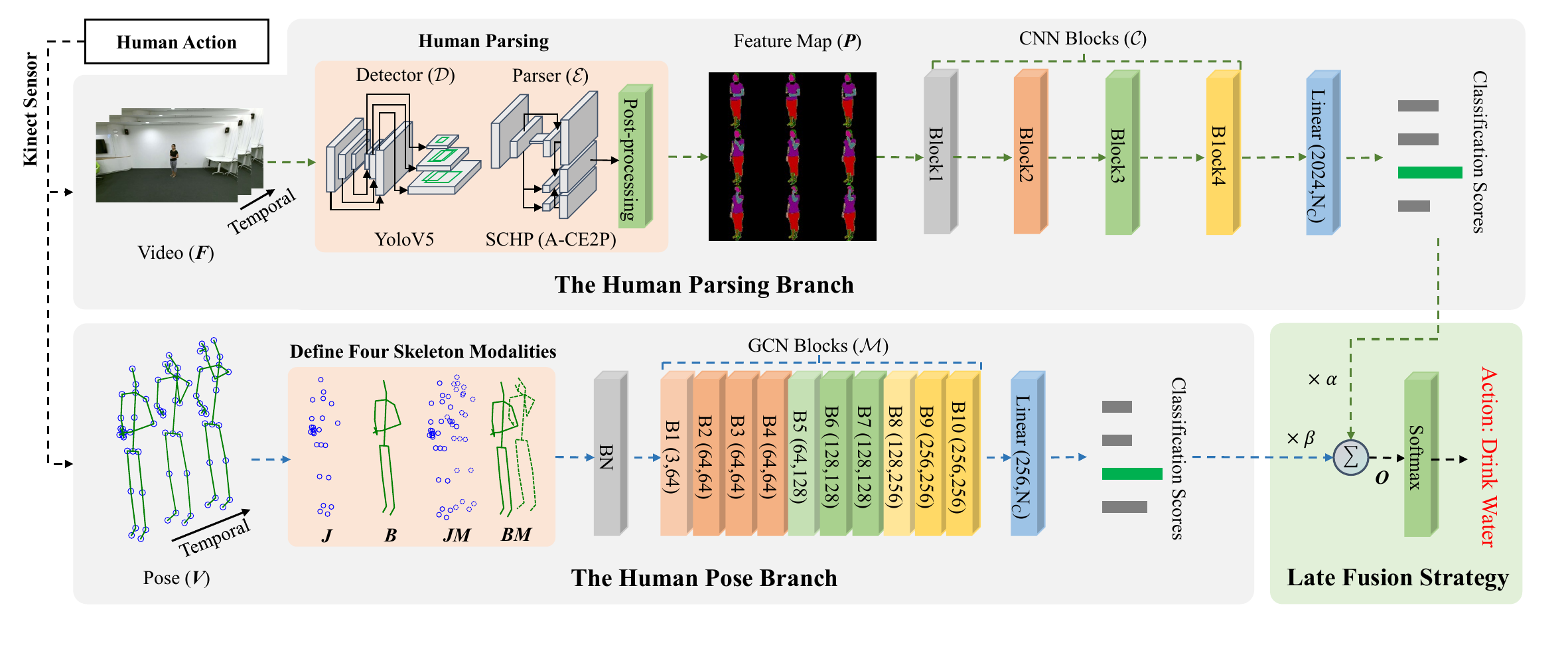}}
\caption{Framework of our proposed Ensemble Human Parsing and Pose Network.}
\label{EPPNet}
\end{figure*}

In this work, we advocate integrating human parsing feature maps as a novel modality into multimodal-based action recognition and propose an Ensemble Human Parsing and Pose Network (EPP-Net), as shown in Fig. \ref{EPPNet}. Specifically, our EPP-Net consists of two trunk branches called human pose branch and human parsing branch. In the human pose branch, skeleton data is transformed into various representations. These skeleton representations are then fed into a GCN to obtain high-level features, e.g. classification scores. In the human parsing branch, the human parsing features from different frames are sequentially combined to construct a feature map, which is also subsequently inputted into a CNN to derive high-level features. These high-level features from both branches are effectively integrated via an ensemble layer for final action recognition. By leveraging the proposed EPP-Net, we effectively integrate pose data and human parsing feature maps to achieve better human action recognition.

Our contributions are three-fold, summarized as follows:
\begin{enumerate}
    \item We advocate leveraging human parsing feature maps as a novel modality for multimodal-based action recognition, which is depictive and can filter out most irrelevant noise. This introduced human parsing modality can be effectively combined with skeleton data for better action recognition.
    \item We propose a new dual-branch framework called Ensemble Human Parsing and Pose Network (EPP-Net), which effectively combines human parsing feature map and pose modality for the first time. In detail, the skeleton data and feature maps are inputted to GCN and CNN backbones for modeling high-level features, which will be effectively combined through a late fusion strategy for more robust action recognition.
    \item Extensive experiments on benchmark NTU RGB+D and NTU RGB+D 120 datasets verify the effectiveness of our EPP-Net. In these two large-scale action recognition datasets, our model outperforms most existing unimodal and multimodal methods.
\end{enumerate}

\section{RELATED WORK}\label{sec2}
\subsection{Human action recognition}
Human action recognition plays a pivotal role in video understanding,  where the human actions convey essential information such as emotions and potential intentions, enabling a deeper understanding of individuals within video content. Initially, researchers used hand-crafted features\cite{Veeriah_2015_ICCV, wang2012mining, 6239233} for action recognition. With the rise of deep learning, more researchers turned their attention to neural networks, giving birth to many classical methods based on CNNs\cite{LIU2017346}, RNNs\cite{Wang_2017_CVPR} and GCNs\cite{shiftgcn2020,dynamicgcn2020,CTR-GCN2021,gsgcn2022,TDGCN,yan2018spatial}. Meanwhile, action recognition has expanded from unimodal-based methods to multimodal-based methods, incorporating modalities such as skeletons\cite{yan2018spatial, CTR-GCN2021}, RGB images\cite{vpn2020, Yu_Liu_Chan_2021, STAR}, point clouds\cite{fan2021pstnet}, depth images\cite{9422825} and human parsing. Currently, achieving robust action recognition through multimodal fusion has gained significant attention. In this section, we will first introduce some unimodal-based and multimodal-based methods in action recognition. Then, we will emphasize introducing some knowledge about human parsing and incorporate the human parsing modality into action recognition. Finally, we will introduce some classical methods for multimodal fusion in human action recognition.

\subsection{Unimodal action recognition}
Action recognition methods can be roughly divided into two ways: unimodal and multimodal. Usually the unimodal methods use a single modality such as pose\cite{dstanet2020,CTR-GCN2021,Chi_2022_CVPR,gsgcn2022,psumnet2023,TDGCN}, RGB images\cite{li2022uniformer,liu2021video} or point clouds\cite{fan2021pstnet} as input. Among them, the skeletons is the most popular which is often used for skeleton-based action recognition. Due to the graph-structured skeleton can well represent body movements and is highly robust to environmental changes, many studies\cite{shiftgcn2020,dynamicgcn2020,CTR-GCN2021,gsgcn2022,TDGCN,yan2018spatial} have used it on GCNs and achieved high success. For example, Liu et al.\cite{TDGCN} proposed a Temporal Decoupling Graph Convolutional Network (TD-GCN), which learn high-level features from skeleton data effectively via using temporal-dependent adjacency matrix. Cheng et al.\cite{shiftgcn2020} proposed a shift-GCN, which used a simple shift operation to swap spatiotemporal features between skeletons. Chen et al.\cite{CTR-GCN2021} proposed a CTR-GCN, which focuses on using channel features to model skeleton dependencies in unimodal action recognition. These unimodal methods usually focus on aggregating unimodal features from skeletons. However, unimodality has inherent limitations in representing human actions. For instance, the skeleton modality lacks appearance of human body and the RGB modality is easily influenced by various noise\cite{SurveyTIT,LIU201614}, while the unordered point clouds lack fine-grained features. These flaws greatly limit the performance of unimodal-based methods and hinder the improvement of action recognition.

\subsection{Multimodal action recognition}
Benefiting from the emergence of various multimodal datasets and improvements in computing resources, multimodal research involving representation, translation, alignment and fusion tasks\cite{8269806} has become popular\cite{ORMF,MCGEC,ST-SIGMA}. Commonly used modalities for multimodal-based action recognition include skeletons\cite{CTR-GCN2021,Joze_2020_CVPR}, color images\cite{tu2023dtcm, vpn2020,9680676,Joze_2020_CVPR}, depth maps\cite{9422825}, text\cite{LST2022,LA-GCN} and point clouds\cite{fan2021pstnet}. For example, Shu et al.\cite{9680676} propose a novel multimodal fusion network called ESE-FN to aggregate discriminative information of skeletons and color images for better action recognition. To better use text modality to assist skeleton-based action recognition, Xu et al.\cite{LA-GCN} leverage large language models (LLM) to generate text features to provide global prior knowledge for skeleton joints. Wu et al.\cite{9422825} use 3D CNN to effectively fuse and complement the skeleton and depth information for robust multimodal action recognition. These multimodal-based methods take full advantage of the complementarity between modalities. However, they still suffer from the inherent imperfections of single modality.

Human parsing task is an important vision task that holds significant importance in video surveillance and human behavior analysis. Human parsing involves the segmentation of a human image into fine-grained semantic parts, including the head, torso, arms, and legs. Several benchmarks have been proposed for human parsing task, providing large-scale annotations of body parts\cite{Chen_2014_CVPR,ATR,LIP2019}. A number of works concentrated on this problem and proposed novel models for better semantic parsing. The majority are based on ResNet architecture\cite{LIP2019,ruan2019devil,Zhao_2017_CVPR,li2020self}, while some are based on HRNet architecture\cite{10.1007/978-3-030-58539-6_11} and Transformer architecture \cite{Chen_2023_CVPR}. Inspired by the human parsing task, we exploit the advantages of human parsing to get noiseless and concise representations, thus introducing the human parsing feature map into multimodal action recognition. Based on the novel human parsing modality, we propose an Ensemble Human Parsing and Pose Network (EPP-Net) for multimodal-based action recognition. Our EPP-Net argues to utilize both pose data and feature map sequences of human parsing. Compared with the above approaches, the human parsing module filters out irrelevant information regarding illumination and backgrounds, while selectively retaining spatiotemporal features of all body parts. This novel modality is proven effective in our ensemble framework. 

Generally the fusion strategies among modalities in multimodal action recognition can be summarized into early fusion\cite{vpn2020}, intermediate fusion\cite{Joze_2020_CVPR} and late fusion\cite{9422825}. Early fusion integrates features immediately after they are extracted, while late fusion performs integration after each modality has made a decision. Intermediate fusion usually uses intermediate level features of each modality for multimodal interactions\cite{Joze_2020_CVPR}. These fusion strategies are widely used in multimodal-based action recognition. For example, Joze et al.\cite{Joze_2020_CVPR} introduced a Multimodal Transfer Module (MMTM) into action recognition, which operates between convolutional neural networks and uses multimodal information to recalibrate features of different modality based on the intermediate fusion strategies. Wu et al.\cite{9422825} proposed a multimodal two-stream 3D network framework, which fuses the classification scores of RGB and depth stream based on the late fusion strategy. Among these three common fusion strategies, late fusion tends to be efficient and concise, and there is no need to design complex fusion modules when using complementary modalities. In our proposed EPP-Net, we also use a late fusion strategy to integrate the classification scores of both human pose and human parsing modalities.

\section{METHOD}\label{sec3}
\subsection{Define four skeleton modalities}
In our proposed EPP-Net, the human pose branch leverages human skeleton data collected by sensors. Conceptually, the skeleton sequence forms a natural topological graph, in which joints are graph vertices, and bones are edges. The graph can be defined as $\textit{\textbf{G}} = \left\{\textit{\textbf{V}}, \textit{\textbf{E}} \right\}$, where $\textit{\textbf{E}} = \left\{\textit{\textbf{e}}_{1}, \textit{\textbf{e}}_{2}, \cdots, \textit{\textbf{e}}_{N} \right\}$ and $\textit{\textbf{V}} = \left\{\textit{\textbf{v}}_{1}, \textit{\textbf{v}}_{2}, \cdots, \textit{\textbf{v}}_{N} \right\}$ represent N bones and N joints of human body. The joint $\textit{\textbf{v}}_{i}$ is defined as $\left\{\textit{\textbf{x}}_{i}, \textit{\textbf{y}}_{i}, \textit{\textbf{z}}_{i} \right\}$ in 3D pose data, where $\textit{\textbf{x}}_{i}$, $\textit{\textbf{y}}_{i}$ and $\textit{\textbf{z}}_{i}$ locate $\textit{\textbf{v}}_{i}$ in three-dimensional Euclidean space.

Here we define skeleton data as four different modalities, namely joint $(\textit{\textbf{J}})$, bone $(\textit{\textbf{B}})$, joint motion $(\textit{\textbf{JM}})$ and bone motion $(\textit{\textbf{BM}})$. Given two joints data $\textit{\textbf{v}}_{i} = \left\{\textit{\textbf{x}}_{i}, \textit{\textbf{y}}_{i}, \textit{\textbf{z}}_{i} \right\}$ and $\textit{\textbf{v}}_{j} = \left\{\textit{\textbf{x}}_{j}, \textit{\textbf{y}}_{j}, \textit{\textbf{z}}_{j} \right\}$, a bone data of the skeleton is defined as a vector $\textit{\textbf{e}}_{\textit{\textbf{v}}_{i}, \textit{\textbf{v}}_{j}} = \left(\textit{\textbf{x}}_{i}-\textit{\textbf{x}}_{j}, \textit{\textbf{y}}_{i}-\textit{\textbf{y}}_{j}, \textit{\textbf{z}}_{i}-\textit{\textbf{z}}_{j} \right)$. Given two joints data $\textit{\textbf{v}}_{ti}$, $\textit{\textbf{v}}_{(t+1)i}$ from two consecutive frames, the data of joint motion is defined as $\textit{\textbf{m}}_{ti} = \textit{\textbf{v}}_{(t+1)i} - \textit{\textbf{v}}_{ti}$. Similarly, given two bones data $\textit{\textbf{e}}_{\textit{\textbf{v}}_{(t+1)i},\textit{\textbf{v}}_{(t+1)j}}$, $\textit{\textbf{e}}_{\textit{\textbf{v}}_{ti},\textit{\textbf{v}}_{tj}}$ from two consecutive frames, the data of bone motion is defined as $\textit{\textbf{m}}_{\textit{\textbf{v}}_{ti},\textit{\textbf{v}}_{tj}} = \textit{\textbf{e}}_{\textit{\textbf{v}}_{(t+1)i},\textit{\textbf{v}}_{(t+1)j}} - \textit{\textbf{e}}_{\textit{\textbf{v}}_{ti},\textit{\textbf{v}}_{tj}}$.

\begin{figure}[t]
\centerline{\includegraphics[width=8cm]{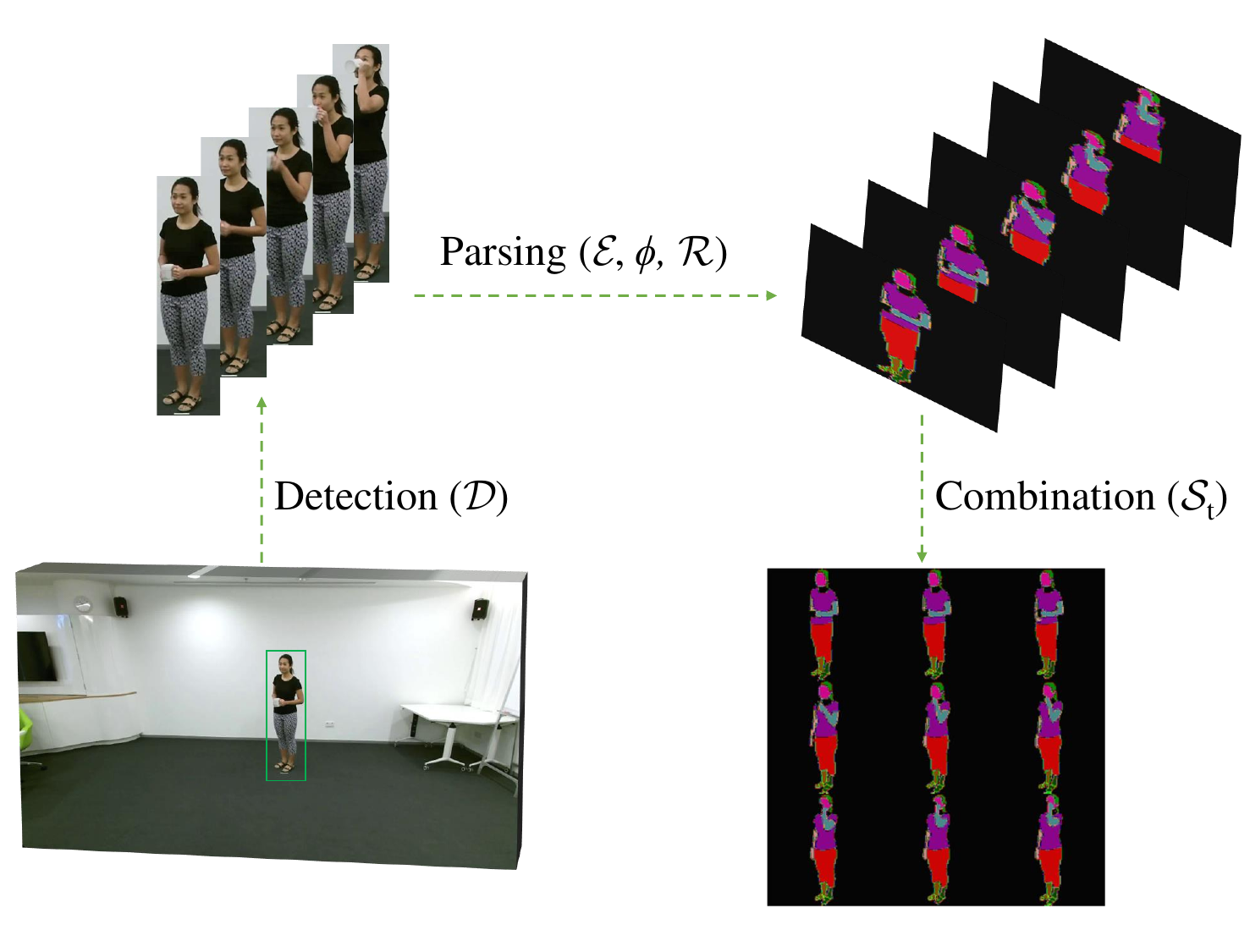}}
\caption{Generate noiseless human parsing feature map from noisy RGB videos.}
\label{parsing}
\end{figure}

\subsection{Generate human parsing feature map}
In our EPP-Net, the human parsing branch leverages the noiseless feature maps as the input modality. These feature maps are generated from noisy RGB videos via the human parsing models based on the following steps, which is shown in Fig. \ref{parsing}. Given an RGB video stream $\textit{\textbf{F}} = \left\{\textit{\textbf{f}}_{1}, \textit{\textbf{f}}_{2}, \cdots, \textit{\textbf{f}}_{N} \right\}$ with N frames, we first extract the features of each frame to filter most noise and irrelevant factors by
\begin{center}
\begin{equation}
\setlength{\abovedisplayskip}{-10pt}
\setlength{\belowdisplayskip}{0pt}
\textit{$\textit{\textbf{I}}_{i}$} = \mathcal E[\mathcal D(\textit{\textbf{f}}_{i})], i \in 1, 2, \ldots, N,
\label{con:Equation2}
\end{equation}
\end{center}
where $\mathcal D$ and $\mathcal E$ denote an object detector and a feature extractor respectively. Inspired by the top-down pose estimation, we use an object detector $\mathcal D$ to filter out most ambient noise and focus on the human body. The feature extractor $\mathcal E$ performs fine-grained human parsing from the detected human anchors to obtain the typical features of body parts. In the implementation, we use YoloV5\cite{yolov5} and SCHP (A-CE2P)\cite{li2020self} as the object detector and human parser to extract useful human parsing features from RGB videos. 

In order to save computing and storage resources, we resize the extracted frame features $\textit{\textbf{I}}_{i}$ and select t frames to construct the final feature map by equation \ref{con:Equation3}.

\begin{center}
\begin{equation}
\setlength{\abovedisplayskip}{-10pt}
\setlength{\belowdisplayskip}{0pt}
\textit{$\textit{\textbf{P}}$} = \mathcal S_{t}[\phi(\mathcal R(\textit{$\textit{\textbf{I}}_{1}$}), \mathcal R(\textit{$\textit{\textbf{I}}_{2}$}), \ldots, \mathcal R(\textit{$\textit{\textbf{I}}_{N}$}))], 1 \le t \le N.
\label{con:Equation3}
\end{equation}
\end{center}

\begin{figure*}[t]
\centerline{\includegraphics[width=16cm]{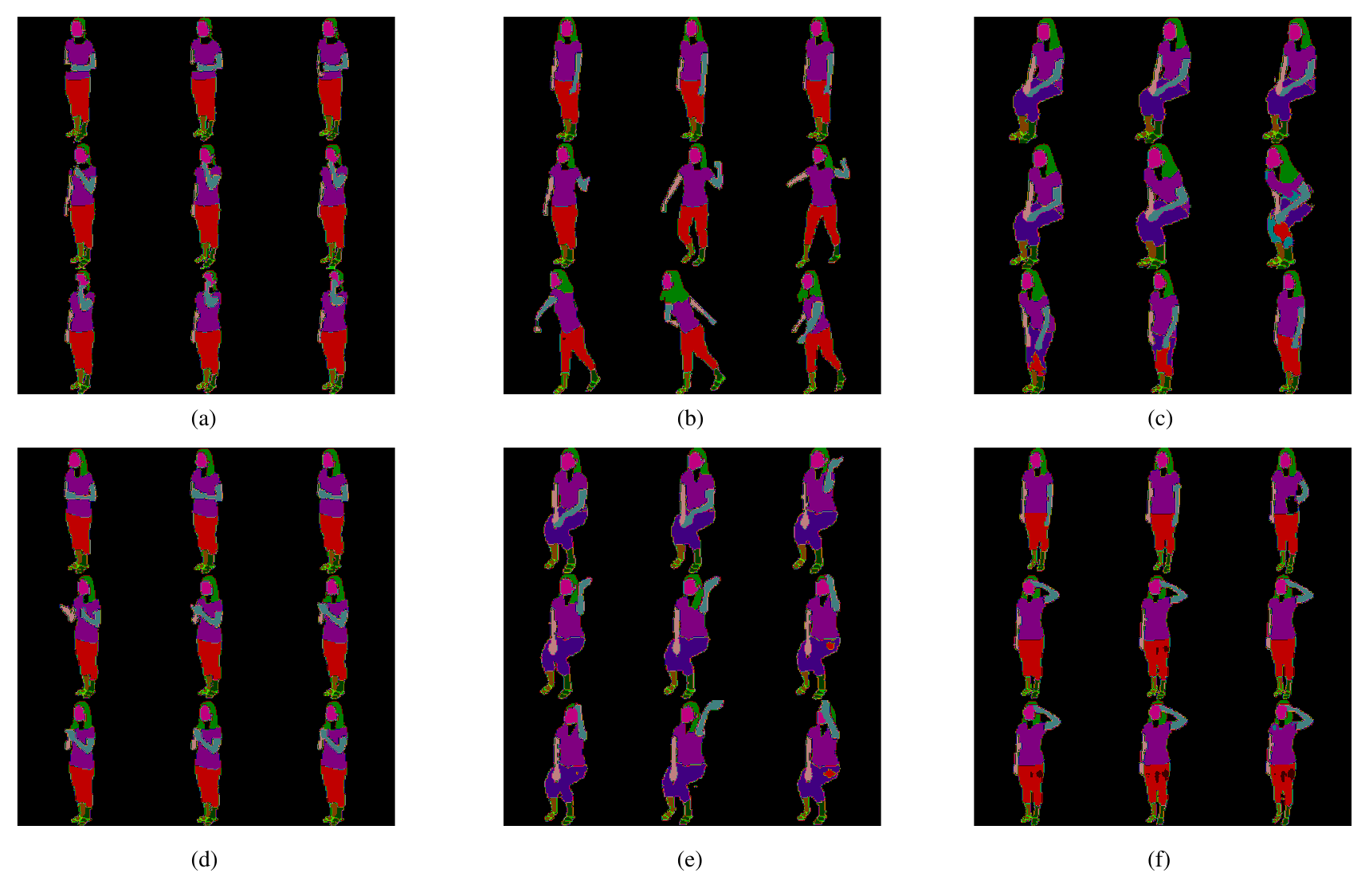}}
\caption{Visualization of human parsing feature maps for different action samples. (a) Feature map of \textit{drink water}. (b) Feature map of \textit{throwing}. (c) Feature map of \textit{stand up}. (d) Feature map of \textit{clapping}. (e) Feature map of \textit{hand waving}. (f) Feature map of \textit{salute}.}
\label{map}
\end{figure*}

In equation \ref{con:Equation3}, the operation $\mathcal R$ resize each frame features to the specified size ($H_s$, $W_s$) and t frames of which will be arranged in chronological order to form the final feature map with a size of ($H_m$, $W_m$) via the function $\mathcal S_{t}$. The operator $\phi$ colorizes the feature by assigning a unique RGB value in color domain to each parsing category. This colorization step magnifies the inter-category difference in features between semantic parts of the human body. It also provides features close to natural RGB image for better vision backbone learning. Fig.~\ref{map} visualizes the feature maps of six action samples, namely \textit{drink water}, \textit{throwing}, \textit{stand up}, \textit{clapping}, \textit{hand waving} and \textit{salute}.

\subsection{Compute unimodal classification scores}
Inspired by GCN's unique advantages in modeling graph-structured data, our EPP-Net uses GCN to obtain skeleton-based unimodal classification scores. Actually the Graph Convolutional Networks consist of two parts: graph convolution module and temporal convolution module. The normal graph convolution can be formulated as:
\begin{equation}
  H^{l+1} = \sigma\left(D^{-\frac{1}{2}}AD^{-\frac{1}{2}}H^{l}W^{l} \right)
  \label{eq:Equation4},
\end{equation}
where $H^{l}$ is the joint features at layer $\textit{l}$ and $\sigma$ is the activation function. $\textit{D} \in \textit{R}^{N \times N}$ is the degree matrix of $\textit{N}$ joints and $\textit{W}^{\textit{l}}$ is the learnable parameter of the $\textit{l}$-th layer. $\textit{A}$ is the adjacency matrix representing joint connections. Generally, the $\textit{A}$ can be generated by using static and dynamic ways. The $\textit{A}$ is generated by using data-driven strategies in dynamic ways while it is defined manually in static ways. In our EPP-Net, we use ten dynamic GCN blocks. The four different skeleton modalities mentioned above will first go through a data normalization layer, then input into the ten GCN blocks for feature extraction. Finally, a linear layer is used to obtain the unimodal classification score.

Convolutional Neural Networks (CNNs) have unique advantages in modeling Euclidean data (e.g. images), which can effectively learn local fine-grained features through the convolution kernel. These convolution layers with specific kernels sizes can further learn high-level features when applied to the body parts of human parsing. Therefore, we use four CNN blocks to process the human parsing feature maps and obtain high-level features for classification. Finally, a linear layer is also used to obtain the unimodal classification score based on the human parsing modality.

\begin{center}
\begin{table*}[t]
\caption{Accuracy comparison with state-of-the-art methods on NTU-RGB+D and NTU-RGB+D 120 dataset.}
\begin{tabular*}{\textwidth}{@{\extracolsep\fill}lllllll@{}}
\toprule
\multirow{2}{*}{\textbf{Modality}} & \multirow{2}{*}{\textbf{Method}} & \multirow{2}{*}{\textbf{Source}} & \multicolumn{2}{c}{\textbf{NTU 60 ($\%$)}} & \multicolumn{2}{c}{\textbf{NTU 120($\%$)}} \\
\cmidrule{4-5}\cmidrule{6-7}
 &  &  & {\textbf{X-Sub}}  & \textbf{X-View} & \textbf{X-Sub} & \textbf{X-Set} \\
\midrule
Skeleton & Shift-GCN \cite{shiftgcn2020} & CVPR'20 & 90.7 & 96.5 & 85.9 & 87.6 \\
Skeleton & DynamicGCN \cite{dynamicgcn2020} & MM'20 & 91.5 & 96.0 & 87.3 & 88.6 \\
Skeleton & DSTA-Net \cite{dstanet2020} & ACCV'20 & 91.5 & 96.4 & 86.6 & 89.0 \\
Skeleton & MS-G3D \cite{MS-G3D2020} & CVPR'20 & 91.5 & 96.2 & 86.9 & 88.4 \\
Skeleton & MST-GCN \cite{chen2021multi} & AAAI'21 & 91.5 & 96.6 & 87.5 & 88.8 \\
Skeleton & CTR-GCN \cite{CTR-GCN2021} & ICCV'21 & 92.4 & 96.8 & 88.9 & 90.6 \\
Skeleton & GS-GCN \cite{gsgcn2022} & CICAI'22 & 90.2 & 95.2 & 84.9 & 87.1 \\
Skeleton & PSUMNet \cite{psumnet2023} & ECCV'22 & 92.9 & 96.7 & 89.4 & 90.6 \\
Skeleton & InfoGCN \cite{Chi_2022_CVPR} & CVPR'22 & 93.0 & 97.1 & 89.8 & 91.2 \\
Skeleton & STSA-Net \cite{STSAnet2023} & {Neuro.'23} & 92.7 & 96.7 & 88.5 & 90.7 \\
\midrule
Skeleton+RGB & VPN \cite{vpn2020} & ECCV'20 & 93.5 & 96.2 & 86.3 & 87.8 \\
Skeleton+RGB & TSMF \cite{Yu_Liu_Chan_2021} & AAAI'21 & 92.5 & 97.4 & 87.0 & 89.1 \\
Depth+RGB & DRDIS \cite{9422825} & TCSVT'22 & 91.1 & 94.3 & 81.3 & 83.4 \\
Skeleton+Text & LST \cite{LST2022} & ICCV'23 & 92.9 & 97.0 & 89.9 & 91.1 \\
Skeleton+RGB & STAR-Transformer \cite{STAR} & WACV'23 & 92.0 & 96.5 & 90.3 & 92.7 \\
Skeleton+Text & LA-GCN \cite{LA-GCN} & arXiv'23 & 93.5 & 97.2 & 90.7 & 91.8 \\
Skeleton+Parsing & Ours (J+B+P) &  & 94.6 & 97.6 & 90.9 & 92.7 \\
\textbf{Skeleton+Parsing} & \textbf{Ours} &  & \textbf{94.7} & \textbf{97.7} & \textbf{91.1} & \textbf{92.8} \\
\bottomrule
\label{tab:sota}
\end{tabular*}
\end{table*}
\end{center}
\vspace{-1cm}
In our proposed EPP-Net, we use Cross-Entropy (CE) loss as the classification loss: 
\begin{equation}
  \mathcal L_{cls} = -\sum_{i}^{N}\mathcal Y^{i}log\mathcal S_{\mathcal M}^i,
  \label{eq:Equation5}
\end{equation}
where $\textit N$ is the number of samples in a batch and $\mathcal Y^{i}$ is the one-hot presentation of the true label about action sample $\textit i$. The $\mathcal S_{\mathcal M}^i$ is the true unimodal classification score of action $\textit i$ output by the linear layer.

\subsection{Late fusion strategy}
Finally, a late fusion strategy is used to better fuse the classification scores of two modalities. Generally the late fusion is efficient and concise when using complementary modalities like pose and human parsing feature maps. In detail, the classification scores from human pose branch and human parsing branch are fused via an ensemble layer, which is formulated as:
\begin{center}
\begin{equation}
\setlength{\abovedisplayskip}{-10pt}
\setlength{\belowdisplayskip}{0pt}
\textit{$\textit{\textbf{O}}$} = \alpha \cdot \mathcal C(\textit{\textbf{P}}) + \beta \cdot \mathcal M(\textit{\textbf{V}}),
\label{con:Equation6}
\end{equation}
\end{center}
where \textit{\textbf{P}} and \textit{\textbf{V}} denote human parsing feature maps and the human pose data respectively. To obtain the classification scores of each modality, we use a CNN $\mathcal C$ and GCN $\mathcal M$ to process the feature maps and skeleton data respectively. Subsequently, the classification scores of two modalities are effectively fused by using two ensemble rates $\alpha$ and $\beta$. The final result \textit{\textbf{O}} that integrates two modalities will be processed by a softmax layer for final action recognition.

\section{EXPERIMENTS}\label{sec4}
\subsection{Datasets}
We use two widely-used large-scale human action recognition datasets in our experiments: \textbf{NTU-RGB+D} (NTU 60) and \textbf{NTU-RGB+D 120} (NTU 120).

\textbf{NTU-RGB+D}\cite{7780484} is a widely used 3D action recognition dataset containing 56,880 skeleton sequences and human action videos, which are categorized into 60 action classes and performed by 40 different performers. The original paper\cite{7780484} suggests two benchmark scenarios for evaluation: (1) Cross-View (X-View), where the training data originates from cameras at $0^\circ$ (view 2) and $45^\circ$ (view 3), and the testing data is sourced from the camera at $45^\circ$ (view 1). (2) Cross-Subject (X-Sub), where the training data comprises samples from 20 subjects, while the remaining 20 subjects are reserved for testing.

\textbf{NTU-RGB+D 120}\cite{8713892} is derived from the NTU-RGB+D dataset. A total of 114,480 video samples across 120 classes are performed by 106 volunteers and captured using three Kinect V2 cameras. The original work\cite{8713892} also suggests two criteria: (1) Cross-subject (X-Sub), where the training data is sourced from 53 subjects, while the testing data originates from the other 53 subjects. (2) Cross-setup (X-Set), where the training data is composed of samples with even setup IDs, and the testing data comprises samples with odd setup IDs.

\subsection{Implementation details}
All experiments are conducted on two NVIDIA GeForce RTX 3070 GPUs and four Tesla V100-PCIE-32GB GPUs. On the human pose branch, we adopt CTR-GCN\cite{CTR-GCN2021} as the backbone to obtain pose-based classification scores. We use SGD for model training, conducting 65 epochs with a batch size of 64. The initial learning rate is set to 0.1 and is decayed by a factor of 0.1 at epochs 35 and 55. On the human parsing branch, we adopt InceptionV3\cite{szegedy2016rethinking} as the backbone to obtain parsing-based classification scores. We also use SGD to train the model for 45 epochs with a batch size of 64. The initial learning rate is set to 0.03 and is decayed by a factor of 0.0001 every 10 epochs. During training we select 9 frames randomly to construct the feature maps of human parsing and randomly change brightness, contrast, saturation for data augmentation, while 9 frames at equal intervals in testing. When generating the human parsing feature map, the human parsing map of each frame will be resized to the specified size of [160, 160] and 9 frames of which will be arranged in chronological order to form the final feature map with a size of [480, 480]. Specifically, we fused the classification scores of joint, bone, joint motion, bone motion and human parsing in a ratio of 2:2:1:1:2.

\begin{figure*}[t]
\centerline{\includegraphics[width=17cm]{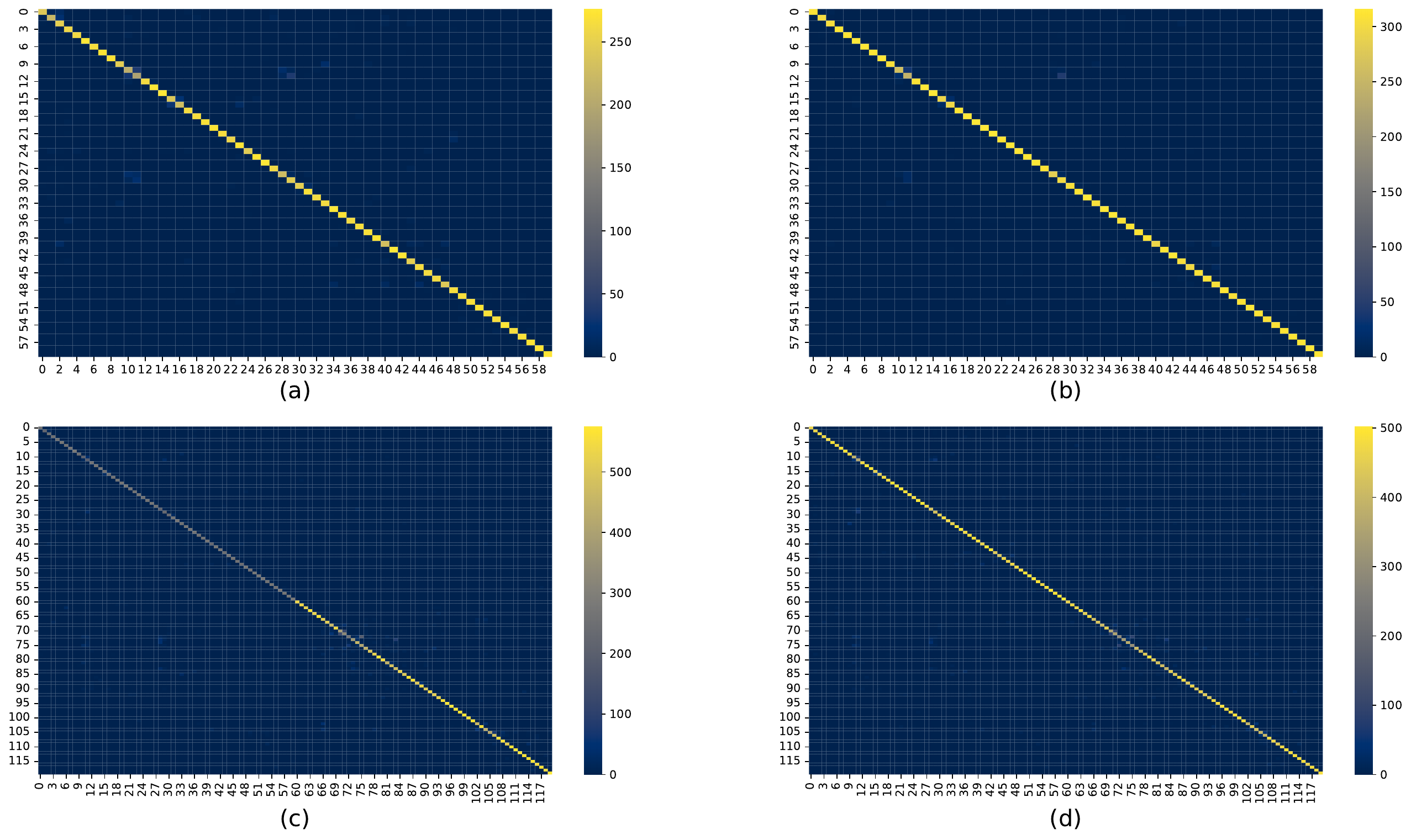}}
\caption{The confusion matrix of NTU60 dataset and NTU120 dataset. The more yellow the squares on the diagonal, the more accurate the recognition. (a) NTU60 dataset on the benchmark of X-Sub. (b) NTU60 dataset on the benchmark of X-View. (c) NTU120 dataset on the benchmark of X-Sub. (d) NTU120 dataset on the benchmark of X-Set.}
\label{matrix}
\end{figure*}

\subsection{Comparison with related methods}
In Table \ref{tab:sota}, we report the \textbf{Top-1} accuracy of our EPP-Net and compare with the existing methods on the NTU-RGB+D and NTU-RGB+D 120 datasets. In these two widely-used action recognition datasets, our EPP-Net outperforms most existing unimodal and multimodal methods under the recommended evaluation benchmarks. It is worth emphasizing that our EPP-Net is the first to combine human parsing feature map and pose data for multimodal-based action recognition. 

In the NTU-RGB+D dataset, the classification accuracy is 94.7$\%$ and 97.7$\%$ on the benchmark of X-Sub and X-View respectively, which outperforms InfoGCN\cite{Chi_2022_CVPR} by 1.7$\%$ and 0.6$\%$. On the tougher benchmark namely X-Sub, our EPP-Net outperforms LST\cite{LST2022} and LA-GCN\cite{LA-GCN} by 1.8$\%$ and 1.2$\%$ although the LST and LA-GCN model additionally introduces a new modality of text data. In the NTU-RGB+D 120 dataset, the classification accuracy is 91.1$\%$ and 92.8$\%$ on the benchmark of X-Sub and X-View respectively, which outperforms CTR-GCN by 2.2$\%$ and 2.2$\%$. Like most multimodal methods, VPN\cite{vpn2020} and TSMF\cite{Yu_Liu_Chan_2021} use RGB and skeleton modalities for better action recognition. On two benchmarks of NTU-RGB+D 120 dataset, our EPP-Net outperforms the VPN by 4.8$\%$ and 5.0$\%$, and outperforms the TSMF by 4.1$\%$ and 3.7$\%$. 

We present the confusion matrices of our EPP-Net in Fig. \ref{matrix} and select the most challenging benchmark (i.e., NTU120 X-Sub) for analysis. In the confusion matrix of the NTU120 X-Sub benchmark, a total of 62 samples achieved an accuracy exceeding 95$\%$, accounting for approximately 51.67$\%$ of the total samples. Meanwhile, there are 98 action samples with an accuracy exceeding 85$\%$, accounting for approximately 81.67$\%$.  Among 120 action samples, the 'staggering', 'jump up', 'arm circles', 'take off jacket', 'walking towards' and 'cheers and drink' action samples have the highest recognition accuracy, reaching 100$\%$, while the 'staple book' action sample has the lowest recognition accuracy, just reaching 44.66$\%$.

\subsection{Ablation study}
In Table \ref{tab:Modality}, we present the recognition accuracy of four skeleton modalities and human parsing feature map in NTU60 and NTU120 datasets respectively. We also explore the ensemble accuracy of different skeleton modalities and feature maps. For example, when the joint, bone and human parsing feature maps are combined (J+B+P), the ensemble recognition accuracy of two benchmarks in the NTU60 dataset is 94.6$\%$ and 97.6$\%$, which outperforms the setting that combining four skeleton modalities (J+B+JM+BM) by 2.2$\%$ and 1.1$\%$ respectively. Similarly, the setting (J+B+P) outperforms the setting (J+B+JM+BM) by 1.9$\%$ and 2.2$\%$ on the two benchmarks of the NTU120 dataset respectively. We also observed that when the skeleton modality is combined with the feature map (such as J+P and B+P), it achieves better accuracy than unimodal method, which indicates the feature map is complementary to the skeleton modality and can contribute to the action recognition task.

\begin{table}[t]
\caption{Accuracy of different modalities on NTU-RGB+D and NTU-RGB+D 120 dataset.}
\centering
\begin{tabular}{ccccccccc}
\hline
\multicolumn{5}{c}{\textbf{Modality}} & \multicolumn{2}{c}{\textbf{NTU 60} ($\%$)} & \multicolumn{2}{c}{\textbf{NTU 120} ($\%$)}\\\hline
J & B & JM & BM & P & \textbf{X-Sub} & \textbf{X-View} & \textbf{X-Sub} & \textbf{X-Set} \\\hline
\Checkmark & & & & & 90.2 & 95.0 & 85.0 & 86.7 \\
& \Checkmark & & & & 90.5 & 94.7 & 86.2 & 87.5 \\
& & \Checkmark & & & 88.1 & 93.2 & 81.2 & 83.0 \\
& & & \Checkmark & & 87.3 & 92.0 & 81.7 & 82.9 \\
& & & & \Checkmark & 84.5 & 86.4 & 69.8 & 74.6 \\\hline
\Checkmark & \Checkmark & & & & 92.2 & 96.2 & 88.7 & 90.2 \\
\Checkmark & \Checkmark & \Checkmark & \Checkmark & & 92.4 & 96.5 & 89.0 & 90.5 \\ 
\Checkmark & & & & \Checkmark & 93.8 & 97.0 & 87.9 & 90.6 \\
& \Checkmark & & & \Checkmark & 93.6 & 96.9 & 89.1 & 91.2 \\
\Checkmark & \Checkmark & & & \Checkmark & 94.6 & 97.6 & 90.9 & 92.7 \\\hline 
\Checkmark & \Checkmark & \Checkmark & \Checkmark & \Checkmark & \textbf{94.7} & \textbf{97.7} & \textbf{91.1} & \textbf{92.8} \\\hline
\end{tabular}
\label{tab:Modality}
\end{table}

In Table \ref{tab:Colorization}, we explore the impact of the colorized and uncolorized parsing feature maps on classification accuracy. When using the colorized feature maps, the accuracy of unimodality is 69.8$\%$, which outperforms the uncolorized setting by 0.3$\%$. Our EPP-Net combines the colorized feature maps and skeleton data for multimodal-based action recognition, which is better than using uncolorized parsing feature maps. We argue that the colorization magnifies the inter-category difference in features between semantic parts of the human body, which is helpful for action recognition. 

\begin{table}[t]
\caption{Performance comparison between the colorized and uncolorized parsing feature maps on NTU-RGB+D 120 X-Sub benchmark. We conducted multiple experiments and reported the average results.}
\centering
\begin{tabular}{ccc}
\hline
\multirow{2}{*}{\textbf{Colorization}} & \multicolumn{2}{c}{\textbf{Modality}} \\
\cline{2-3}
& \textbf{Parsing} ($\%$) & \textbf{Ensemble} ($\%$) \\\hline
w/ & \textbf{69.8} & \textbf{91.1} \\
w/o & 69.5 & 90.8 \\\hline
\end{tabular}
\label{tab:Colorization}
\end{table}

\begin{table}[t]
\caption{Performance of various vision backbones on NTU-RGB+D 120 X-Sub benchmark. The Parsing($\%$) represent the accuracy of the human parsing branch when using different vision backbones.}
\centering
\begin{tabular}{lccccc}
\hline
\multirow{2}{*}{\textbf{Backbone}} & \multirow{2}{*}{\textbf{Input Size}} & \multicolumn{2}{c}{\textbf{Modality}}\\
\cline{3-4}
& & \textbf{Parsing} ($\%$) & \textbf{Ensemble} ($\%$)\\\hline
VGG11 \cite{simonyan2015deep} & 224$\times$224 & 64.6 & 90.7 \\
VGG13 \cite{simonyan2015deep} & 224$\times$224 & 65.0 & 90.8 \\
ResNet18 \cite{he2015deep} & 224$\times$224 & 64.6 & 90.6 \\
ResNet34 \cite{he2015deep} & 224$\times$224 & 65.1 & 90.7 \\
ConvnextV2 Tiny \cite{Woo_2023_CVPR} & 224$\times$224 & 66.7 & 90.8 \\
ConvnextV2 Tiny \cite{Woo_2023_CVPR} & 384$\times$384 & 68.9 & 91.0 \\
EVA-02 Small \cite{EVA02} & 224$\times$224 & 62.2 & 90.5 \\
InceptionV3 \cite{szegedy2016rethinking} & 299$\times$299 & \textbf{69.8} & \textbf{91.1} \\\hline
\end{tabular}
\label{tab:Backbone}
\end{table}

\begin{table}[t]
\caption{Performance of various human parsers on NTU-RGB+D 120 datasets.}
\centering
\begin{tabular}{cccccc}
\hline
\multirow{2}{*}{\textbf{Modality}} & \multirow{2}{*}{\textbf{Human Parser}} & \multicolumn{2}{c}{\textbf{NTU 120 ($\%$)}}\\ 
\cline{3-4}
& & \textbf{X-Sub} & \textbf{X-Set}\\\hline
skeleton & - & 89.4 & 91.2 \\
parsing & PSP-Net\cite{Zhao_2017_CVPR} & 53.6 & 65.8 \\
skeleton+parsing & PSP-Net\cite{Zhao_2017_CVPR} & 90.0 & 91.7 \\
\textbf{parsing} & \textbf{SCHP (A-CE2P)}\cite{li2020self} & \textbf{69.8} & \textbf{74.6} \\
\textbf{skeleton+parsing} & \textbf{SCHP (A-CE2P)}\cite{li2020self} & \textbf{91.1} & \textbf{92.8} \\\hline
\end{tabular}
\label{tab:humanparser}
\end{table}

\begin{table}[t]
\caption{Comparison of different numbers of frames in feature map construction on NTU-RGB+D 120 X-Sub benchmark.}
\centering
\begin{tabular}{ccc}
\hline
\multirow{2}{*}{\textbf{$\#$Frame}} & \multicolumn{2}{c}{\textbf{Modality}} \\
\cline{2-3}
& \textbf{Parsing} ($\%$) & \textbf{Ensemble} ($\%$) \\\hline
4 & 62.0 & 90.9 \\
9 & 69.8 & \textbf{91.1}  \\
16 & 72.6 & 91.0 \\
25 & 71.4 & 91.0 \\\hline
\end{tabular}
\label{tab:Frame}
\end{table}

\begin{table}[t]
\caption{Comparison of parameter and computation costs when training and inferring on a single action sample.}
\centering
\begin{tabular}{cccc}
\hline
\textbf{Modality} & \textbf{Method} & \textbf{Param.} & \textbf{Flops} \\\hline
Depth+RGB & \makecell*[c]{DRDIS \cite{9422825}\\\footnotesize{(Res3D-101+Res3D-101)}} & 171.98M & 27.92G \\
Skeleton+RGB & \makecell*[c]{STAR-Transformer \cite{STAR}\\\footnotesize{(ResNet-MC18+Pure-VIT)}} & 58.49M & 18.67G \\
\textbf{Skeleton+Parsing} & \textbf{\makecell*[c]{Ours\\\footnotesize{(CTR-GCN+Inception-V3)}}} & \textbf{25.27M} & \textbf{7.84G} \\
\hline
\end{tabular}
\label{tab:Flops}
\end{table}

\begin{table}[t]
\caption{Comparison of Classification Scores when using different Modalities. The table shows the classification score of the ground truth category with values ranging from 0.0 to 1.0.}
\centering
\begin{tabular}{cccc}
\hline
\multirow{2}{*}{\textbf{Action Samples}} & \multicolumn{3}{c}{\textbf{Classification Scores}} \\
& Pose & Parsing & Pose+Parsing \\\hline
\makecell*[c]{Type on a keyboard \\\footnotesize{(S001C001P007R002A030)}} & 0.9892 & 0.0060 & 0.9980 \\
\makecell*[c]{Cutting nails \\\footnotesize{(S027C001P006R002A075)}} & 0.0005 & 0.8720 & 0.9990 \\
\makecell*[c]{Yawn \\\footnotesize{(S027C001P006R001A103)}} & 0.0456 & 0.9709 & 0.9999 \\
\makecell*[c]{Blow nose \\\footnotesize{(S027C003P006R001A105)}} & 0.8674 & 0.0931 & 0.9999 \\
\hline
\end{tabular}
\label{tab:scores}
\end{table}

Table \ref{tab:Backbone} shows the impact of using different CNN backbones to process human parsing feature maps on recognition accuracy. Among them, the setting using InceptionV3\cite{szegedy2016rethinking} achieved the best recognition accuracy, which outperforms the VGG13\cite{simonyan2015deep} and ResNet34\cite{he2015deep} by 4.8$\%$ and 4.7$\%$ in unimodal accuracy. Table \ref{tab:humanparser} shows the impact of using different human parser to generate human parsing feature maps on recognition accuracy. In Table \ref{tab:Frame}, we also explore the impact of using different frames to construct feature maps on recognition accuracy. To avoid lack of information or redundancy, our proposed EPP-Net selects 9 frames feature to construct feature maps, which achieves 91.1$\%$ recognition accuracy in the X-Sub benchmark of NTU120 dataset. Additionally, we showcase the parameter and computation cost required for the proposed EPP-Net in Table \ref{tab:Flops}.

Our observation is three-fold based on the experimental results in Table \ref{tab:Modality} and Table \ref{tab:Backbone}. First, the feature maps show desirable performance across different CNN backbones, which demonstrates that the human parsing modality is robust. Second, better recognition performance is obtained when the feature maps are effectively combined with different skeleton modalities (e.g. joint and bone), which also shows that the human parsing feature map is effective. Third, when the human parsing modality is combined with a single skeleton modality, better recognition accuracy will be obtained than the combination of multiple different skeleton modalities. This phenomenon confirms the human parsing modality can better supplement the skeleton information compared with the homogeneous modality.

\subsection{Visualization}
\begin{figure}[t]
\centerline{\includegraphics[width=8cm]{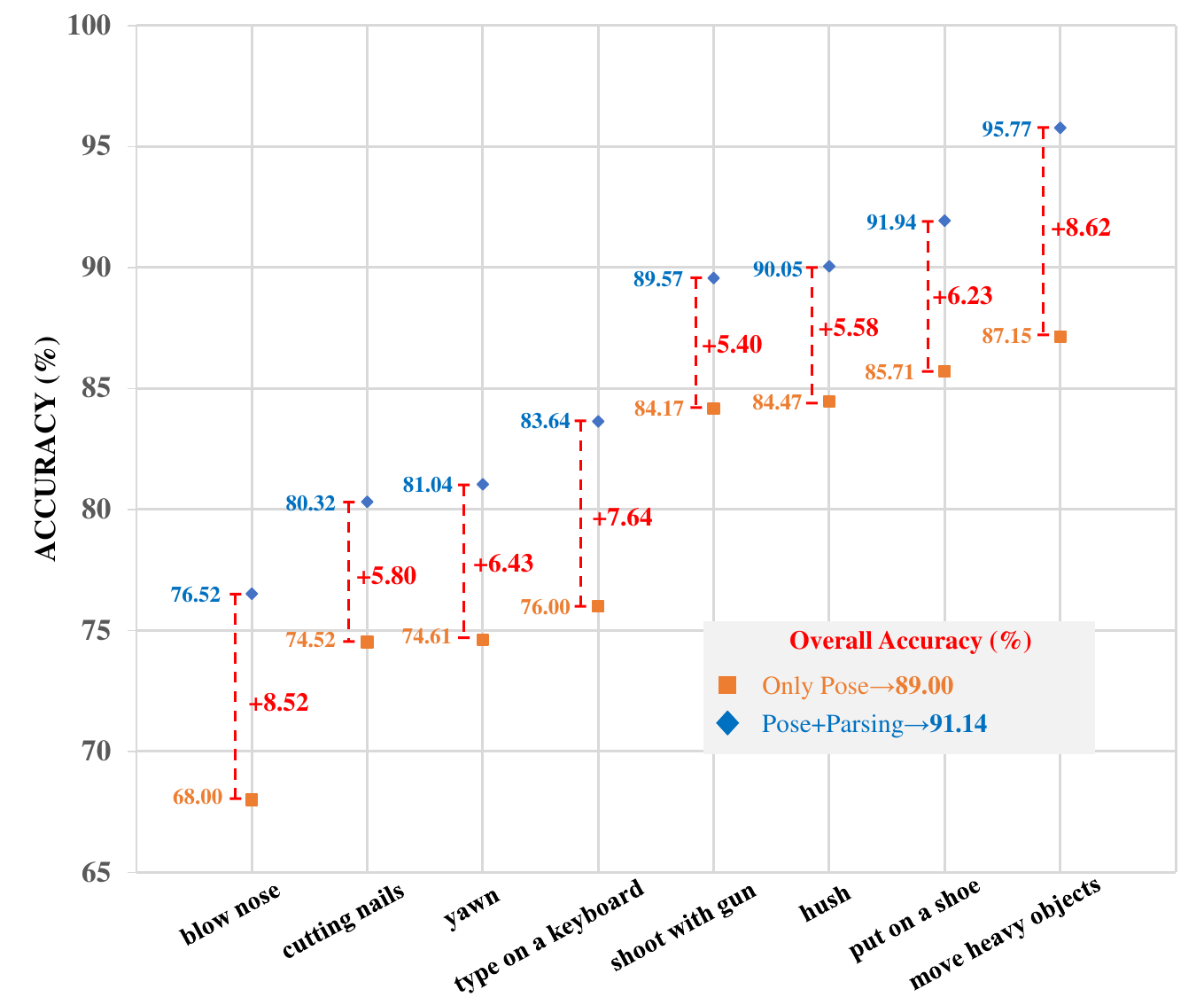}}
\caption{Visualization of the top-8 actions with the highest improvement when the human pose branch is integrated with the human parsing branch.}
\label{vis_action}
\end{figure}

To better illustrate the fusion effect of the human pose branch and human parsing branch, we visualized the overall accuracy of using only pose and the fusion accuracy between two branches on the NTU 120 X-Sub benchmark. We also showcased the top-8 action categories with the highest improvement after adding the human parsing branch as shown in Fig. \ref{vis_action}. Among all actions, the action 'move heavy objects' has the highest improvement, with an accuracy of 87.15$\%$ when using only the human pose branch. Upon fusing the human pose branch and the human parsing branch, the accuracy of the action 'move heavy objects' increases to 95.77$\%$, resulting in an improvement of 8.62$\%$. To further demonstrate the impact of the Human parsing branch, we selected four samples from four action categories in Fig. \ref{vis_action} for analysis, which are shown in Table \ref{tab:scores}. When two complementary modalities (pose and parsing) are fused, it can improve the classification score of the sample in the ground truth category, which substantiate the positive influence of the Human parsing branch.

\begin{figure}[t]
\centerline{\includegraphics[width=8cm]{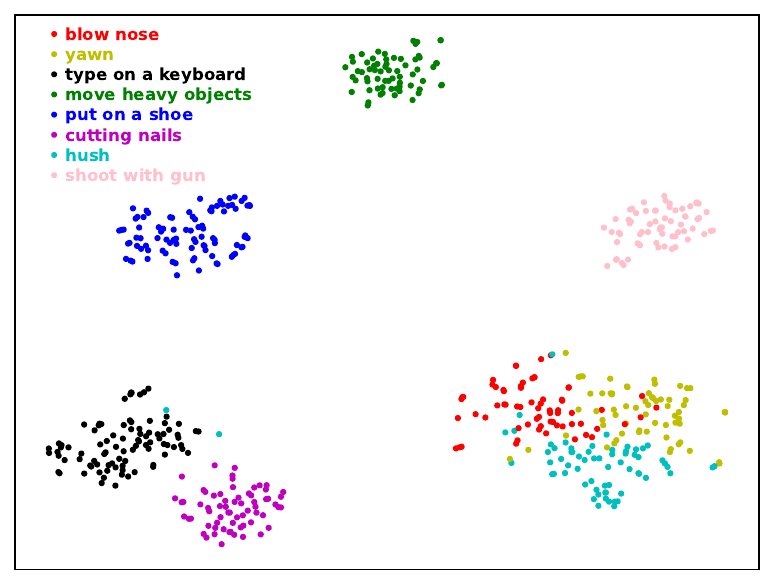}}
\caption{TSNE visualization of skeleton features modeled by GCN for different action categories.}
\label{tsne}
\end{figure}

In addition, in order to better demonstrate the modeling ability and effectiveness of GCN on skeleton data, we performed TSNE visualization of the features about the 8 action categories in Fig. \ref{vis_action} to analyze the relationship between different action categories in the feature space, which is shown in Fig. \ref{tsne}. We observed that the skeleton features of the same category are close and the skeleton features of different categories are far apart in the feature space. Meanwhile, action samples of different categories but with similarities are close in the feature space. For example, the 'blow nose', 'yawn' and 'hush' actions in Fig. \ref{tsne} are very close in the feature space. We argue that this phenomenon is due to the fact that the above three actions mainly occur on the head and face of the human body, leading to their features being close in the feature space.

\section{CONCLUSIONS}\label{sec5}
We propose a new dual-branch framework called Ensemble Human Parsing and Pose Network (EPP-Net) for multimodal-based action recognition, which introduces the depictive and noiseless human parsing feature maps as a novel modality to represent human actions. Different from previous methods using noisy modality, our EPP-Net is the first to leverage both skeletons and human parsing feature maps with the aim of robust action recognition. The effectiveness of EPP-Net is verified on the NTU-RGB+D and NTU-RGB+D 120 datasets, where our EPP-Net outperforms most existing methods.

\bmsection*{Author contributions}
Jinfu Liu and Runwei Ding are co-first authors of this work with equal contributions.

\bibliographystyle{wileyNJD_MPS}
\bibliography{wileyNJD-MPS}

\end{document}